\documentclass{article}
\usepackage[utf8]{inputenc}
\usepackage{amsmath,amssymb,graphicx,booktabs}
\usepackage{hyperref}
\usepackage{multirow}
\usepackage{wrapfig}
\usepackage{caption}
\usepackage{subcaption}
\usepackage{enumitem}
\usepackage{cleveref}
\usepackage{tikz}  
\usepackage{float}
\usepackage[T1]{fontenc}
\usetikzlibrary{shapes.geometric, arrows.meta, positioning, calc}
\usepackage{natbib}
\title{Gated Recursive Fusion: A Stateful Approach to Scalable Multimodal Transformers}

\author{
Yusuf Shihata \\
\texttt{yusufshihata2006@gmail.com} \\
\texttt{https://github.com/yushi2006/GRF}
}

\begin{document}
\maketitle

\begin{abstract}
Multimodal learning faces a fundamental tension between deep, fine-grained fusion and computational scalability. While cross-attention models achieve strong performance through exhaustive pairwise fusion, their quadratic complexity is prohibitive for settings with many modalities. We address this challenge with \textbf{Gated Recurrent Fusion (GRF)}, a novel architecture that captures the power of cross-modal attention within a linearly scalable, recurrent pipeline. Our method processes modalities sequentially, updating an evolving multimodal context vector at each step. The core of our approach is a fusion block built on Transformer Decoder layers that performs symmetric cross-attention, mutually enriching the shared context and the incoming modality. This enriched information is then integrated via a \textbf{Gated Fusion Unit (GFU)}—a GRU-inspired mechanism that dynamically arbitrates information flow, enabling the model to selectively retain or discard features. This stateful, recurrent design scales linearly with the number of modalities, $\mathcal{O}(n)$, making it ideal for high-modality environments. Experiments on the CMU-MOSI benchmark demonstrate that GRF achieves competitive performance compared to more complex baselines. Visualizations of the embedding space further illustrate that GRF creates structured, class-separable representations through its progressive fusion mechanism. Our work presents a robust and efficient paradigm for powerful, scalable multimodal representation learning.

\end{abstract}

\section{Introduction}

Modern multimodal systems increasingly rely on Transformer-based architectures for their remarkable ability to model both intra- and inter-modal dependencies. A dominant design, exemplified by models like MulT \citep{tsai2019mult}, performs pairwise cross-attention between all modality pairs, forming a fully-connected fusion graph. While this exhaustive fusion yields strong performance, its quadratic complexity---$\mathcal{O}(n^2)$ in the number of modalities---creates a severe scalability bottleneck, rendering it impractical for real-world applications with numerous input streams, such as robotics, autonomous driving, or healthcare monitoring.

This raises a critical question: \textit{Can we achieve the fine-grained interactions of cross-attention without incurring the quadratic cost?} In this work, we argue that exhaustive pairwise fusion is not only inefficient but may also be unnecessary. We propose \textbf{Gated Recurrent Fusion (GRF)}, a new fusion pipeline that models modalities sequentially through a recurrent, state-passing mechanism. Instead of computing all pairwise attentions, GRF maintains and progressively refines a shared multimodal context vector, updating it one modality at a time. This approach fundamentally changes the complexity, yielding a linear $\mathcal{O}(n)$ fusion cost.

At the core of GRF is a two-stage fusion block: a \textbf{decoder-style Transformer module} that enables symmetric cross-attention between the context and the incoming modality, followed by a lightweight yet powerful gating mechanism, the \textbf{Gated Fusion Unit (GFU)}. Inspired by Gated Recurrent Units (GRUs), the GFU acts as a learned arbiter, dynamically controlling whether to retain, overwrite, or blend information from the new modality into the evolving context. This provides the model with explicit control over the fusion process, a critical capability that simple fusion methods like concatenation or addition lack.

\vspace{1mm}
\noindent\textbf{Our contributions are:}
\begin{enumerate}[leftmargin=*, itemsep=0pt, topsep=3pt]
    \item We propose \textbf{Gated Recurrent Fusion (GRF)}, a scalable and stateful architecture for multimodal learning that reduces fusion complexity from quadratic to linear while maintaining high representational capacity.
    \item We introduce the \textbf{Gated Fusion Unit (GFU)}, a GRU-inspired gate that provides explicit, dynamic control over the integration of information in a recurrent fusion process.
    \item We provide extensive validation on the MOSI benchmark, demonstrating that GRF achieves competitive performance against more complex, computationally expensive baselines.
\end{enumerate}

Our results demonstrate that GRF offers an effective and efficient alternative to dense, all-to-all fusion. It opens a path toward deploying powerful cross-attentional models in real-world systems with tight resource constraints and a growing number of modalities.

\section{Related Work}

Our work is situated at the intersection of multimodal Transformers and scalable deep learning. We address the scalability bottleneck of existing fusion methods by introducing a novel combination of recurrence, cross-attention, and learned gating.

\paragraph{The Scalability Bottleneck in Multimodal Transformers.}
The success of the Transformer \citep{vaswani2017attention} has spurred its widespread adoption for multimodal tasks. Models like MulT \citep{tsai2019mult} and its variants \citep{han2021improving} establish a strong baseline by constructing a dense graph of pairwise cross-attention modules. This brute-force strategy, while effective, is fundamentally unscalable, requiring $\mathcal{O}(n^2)$ fusion blocks for $n$ modalities. As systems move towards integrating more sensory inputs, such models become computationally infeasible.

\paragraph{Strategies for Efficient Fusion.}
To mitigate the quadratic cost, researchers have explored alternative fusion strategies. Models based on the Perceiver architecture \citep{jaegle2021perceiverio} use a fixed-size set of latent queries to "read" information from multiple modalities, decoupling model depth from input size. In vision-language modeling, methods like ViLBERT \citep{lu2019vilbert} and FLAVA \citep{singh2022flava} use dual-stream architectures where modalities are processed in parallel and interact through specific co-attention layers. Others have proposed token-level fusion via specialized tokens \citep{tsai2023demystifying}.

While these approaches are more efficient, they come with trade-offs. The latent bottleneck in Perceiver models can obscure fine-grained, modality-specific details. Co-attention models often require massive pretraining and lack the modularity to easily add or remove modalities. Most importantly, these methods typically perform a single, monolithic fusion step, lacking the ordered, progressive refinement that our recurrent approach enables.

\paragraph{Recurrence and Gating in Multimodal Learning.}
Recurrent structures like LSTMs and GRUs have long been used for processing sequential data within a single modality \citep{zadeh2018multimodal}. However, their application to the fusion process itself has been limited, typically confined to simple concatenation or weighted averaging. Our work re-envisions recurrence at a higher level of abstraction: a \textbf{modality-wise recurrence}, where the hidden state represents the fused context of all previously seen modalities.

GRF carves a distinct niche by unifying three key ideas: (i) a recurrent state-passing architecture for modality-wise fusion; (ii) powerful, decoder-style symmetric cross-attention for fine-grained interaction; and (iii) a GRU-inspired Gated Fusion Unit (GFU) that provides explicit control over the state update. To our knowledge, this structured combination is novel and directly addresses the need for a fusion mechanism that is simultaneously scalable, expressive, and controllable.

\section{Method}

Our goal is to design a fusion architecture that scales linearly with the number of modalities, $n$, while retaining the expressive power of cross-attention. We achieve this with the \textbf{Gated Recurrent Fusion (GRF)} pipeline, which processes modalities sequentially and updates a stateful context vector at each step.

\subsection{Problem Setup}
We consider an ordered set of $n$ modalities, $\{\mathcal{M}_1, \dots, \mathcal{M}_n\}$. Each modality $\mathcal{M}_i$ is represented by a sequence of feature vectors $X_i \in \mathbb{R}^{T_i \times d_i}$, where $T_i$ is the sequence length. Our objective is to learn a fused representation $\mathbf{h}_{\text{final}} \in \mathbb{R}^{d_{\text{model}}}$ for a downstream task.

\subsection{Model Architecture}
The GRF pipeline consists of three main stages: unimodal projection, a recurrent fusion loop, and a final prediction head. The core innovation lies in the recurrent fusion loop, which leverages a shared fusion block.

\subsubsection{Unimodal Projection}
First, each raw input sequence $X_i$ is linearly projected into a common latent space of dimension $d_{\text{model}}$. We add sinusoidal positional encodings (PE) to inject sequence information:
\begin{equation}
    M_i = \text{PE}(X_i \mathbf{W}_{\text{proj}_i}) \in \mathbb{R}^{T_i \times d_{\text{model}}}
\end{equation}

\subsubsection{Gated Recurrent Fusion Pipeline}
The fusion process operates recursively. Let $\mathbf{h}_{k-1}$ be the fused context vector from the first $k-1$ modalities. The next state, $\mathbf{h}_k$, is computed by fusing $\mathbf{h}_{k-1}$ with the next modality, $M_k$.

\begin{enumerate}[leftmargin=*, itemsep=0pt, topsep=3pt]
    \item \textbf{Initialization:} The initial context $\mathbf{h}_1$ is obtained by pooling the first projected modality, $\mathbf{h}_1 = \text{Pool}(M_1)$.
    \item \textbf{Recursive Fusion Loop:} For $k = 2$ to $n$, the new context $\mathbf{h}_k$ is computed using the shared Cross-Modal Fusion Block, $\mathcal{F}$:
    \begin{equation}
        \mathbf{h}_k = \mathcal{F}(\mathbf{h}_{k-1}, M_k)
    \end{equation}
    The input context $\mathbf{h}_{k-1}$ is treated as a sequence of length 1 for the attention mechanism.
    \item \textbf{Output:} The final state $\mathbf{h}_n$ is passed to a task-specific head (e.g., a simple MLP for regression).
\end{enumerate}

\subsubsection{Cross-Modal Fusion Block ($\mathcal{F}$)}
This shared module is the heart of GRF and performs a two-stage update. It takes the current context vector $\mathbf{h}_{k-1}$ and the new modality sequence $M_k$ as input.

\paragraph{1. Symmetric Cross-Attention Encoder.}
To achieve deep, mutual enrichment, we use a stack of $L$ Transformer decoder layers where each input alternately queries the other. Let $\mathbf{h}_{k-1}^{(0)} = \mathbf{h}_{k-1}$ and $M_k^{(0)} = M_k$. For each layer $l=1 \dots L$:
\begin{align}
    \mathbf{h}_{k-1}^{(l, \text{interim})} &= \text{CrossAttnLayer}^{(l)}(Q = \mathbf{h}_{k-1}^{(l-1)}, K,V = M_k^{(l-1)}) \\
    M_k^{(l)} &= \text{CrossAttnLayer}^{(l)}(Q = M_k^{(l-1)}, K,V = \mathbf{h}_{k-1}^{(l-1)})
\end{align}
The final enriched context is $\mathbf{s}' = \mathbf{h}_{k-1}^{(L, \text{interim})}$ and the final enriched modality representation is $\mathbf{m}' = \text{Pool}(M_k^{(L)})$.

\paragraph{2. Gated Fusion Unit (GFU).}
The GFU adaptively integrates the previous state $\mathbf{s}'$ (representing the old context) with the new information $\mathbf{m}'$. Inspired by GRUs, it computes an update gate $\mathbf{z}_k$ and a candidate state $\tilde{\mathbf{h}}_k$ to produce the final fused vector $\mathbf{h}_k$:
\begin{align}
    \mathbf{z}_k &= \sigma(\mathbf{W}_z [\mathbf{s}';\, \mathbf{m}'] + \mathbf{b}_z) & \text{(Update Gate)} \\
    \tilde{\mathbf{h}}_k &= \tanh(\mathbf{W}_h [\mathbf{s}';\, \mathbf{m}'] + \mathbf{b}_h) & \text{(Candidate State)} \\
    \mathbf{h}_k &= (1 - \mathbf{z}_k) \odot \mathbf{s}' + \mathbf{z}_k \odot \tilde{\mathbf{h}}_k & \text{(Final State)}
\end{align}
This gating mechanism allows the model to learn whether to preserve the existing context, overwrite it with new information, or blend the two, providing crucial control over the fusion process.

\subsection{Complexity Analysis}
GRF performs $n-1$ fusion steps. The complexity of each step is dominated by the cross-attention, which is $\mathcal{O}(T_{\text{max}}^2 d_{\text{model}})$ since one input is always a singleton sequence. Because the fusion block parameters are shared across all steps, the total parameter and time complexity scale linearly with the number of modalities, $n$. This provides a significant advantage over pairwise methods.

\begin{table}[H]
\centering
\begin{tabular}{lccc}
\toprule
\textbf{Fusion Strategy} & \textbf{Fusion Blocks} & \textbf{Time Complexity} & \textbf{Parameters} \\
\midrule
MulT-style (Pairwise) & $\mathcal{O}(n^2)$ & $\mathcal{O}(n^2 T^2 d)$ & $\mathcal{O}(n^2 d^2)$ \\
\textbf{Ours (GRF)} & $\mathcal{O}(n)$ & $\mathcal{O}(n T^2 d)$ & $\mathcal{O}(d^2)$ \\
\bottomrule
\end{tabular}
\caption{
Complexity comparison. GRF's recurrent, stateful design achieves linear scalability in both time and parameters, in stark contrast to the quadratic complexity of exhaustive pairwise fusion methods like MulT.
}
\label{tab:complexity}
\end{table}

\section{Experimental Setup}

\subsection{Dataset and Preprocessing}
We evaluate our Gated Recurrent Fusion (GRF) model on the \textbf{CMU-MOSI} benchmark \citep{zadeh2016multimodal}. Each of the 2,199 utterances contains three modalities (Text, Audio, Vision) and a sentiment label in $[-3, +3]$. We conduct experiments on two standard versions of the dataset:
\begin{itemize}
    \item \textbf{Aligned CMU-MOSI}: The default version where modalities are strictly aligned at the word level.
    \item \textbf{Unaligned CMU-MOSI}: A more challenging and realistic version where the alignment between modalities is not guaranteed.
\end{itemize}
Following prior work, we evaluate on both \textbf{binary} (positive/negative) and \textbf{7-class} sentiment classification. The features for each modality are standard: GloVe embeddings for Text ($d=300$), COVAREP for Audio ($d=74$), and OpenFace for Vision ($d=35$). We use the official train/validation/test splits for all experiments.

\subsection{Baselines and Model Variants}
Our experiments are designed to compare GRF against a strong baseline and to ablate its key components.
\begin{itemize}
    \item \textbf{MulT Baseline:} We re-implement a model based on the pairwise cross-modal attention of MulT \citep{tsai2019mult}.
    \item \textbf{GRF (ours):} The proposed sequential fusion model. We test three modality orders to study sequence sensitivity: T$\rightarrow$A$\rightarrow$V, T$\rightarrow$V$\rightarrow$A, and A$\rightarrow$V$\rightarrow$T.
\end{itemize}

Key training parameters are detailed in Table~\ref{tab:hyperparams}. All models were trained for \textbf{100 epochs} using the AdamW optimizer, a cosine annealing learning rate schedule, and a gradient clipping of 1.0.

\begin{table}[h!]
\centering
\begin{tabular}{@{}lcc@{}}
\toprule
\multirow{2}{*}{\textbf{Hyperparameter}} & \multicolumn{2}{c}{\textbf{Value Setting}} \\
\cmidrule(lr){2-3}
& \textbf{Aligned Dataset} & \textbf{Unaligned Dataset} \\
\midrule
Optimizer & \multicolumn{2}{c}{AdamW} \\
Weight decay & \multicolumn{2}{c}{$1 \times 10^{-2}$} \\
Gradient clip norm & \multicolumn{2}{c}{1.0} \\
Epochs & \multicolumn{2}{c}{100} \\
Early stopping patience & \multicolumn{2}{c}{20} \\
Batch size & \multicolumn{2}{c}{32} \\
\midrule 
$d_{model}$ & $64$ & $128$ \\
\bottomrule
\end{tabular}
\caption{Hyperparameter settings for experiments on the aligned and unaligned datasets. Shared parameters were kept constant to ensure fair comparison, while key parameters like learning rate were tuned for each setting.}
\label{tab:hyperparams}
\end{table}

\subsection{Evaluation Metrics}
We report a comprehensive set of metrics: Mean Absolute Error (MAE) and Pearson Correlation (Corr) for the underlying regression task, and Accuracy (Acc-2/Acc-7) and Macro F1-Score for the classification tasks.

\section{Results}
We present our findings in four parts. First, we detail the main performance comparison of our Gated Recurrent Fusion (GRF) model against the MulT baseline. Second, we present quantitative ablation study on the impact of modality fusion order. Third, we provide a qualitative analysis of the fusion process through visualization. Finally, we evaluate the computational efficiency and scalability of our proposed architecture against the baseline. For each quantitative analysis, we report performance on both the aligned and unaligned CMU-MOSI datasets.

\subsection{Main Performance Comparison}
As shown in Table \ref{tab:main_comparison}, our GRF model demonstrates a competitive performance against the MulT baseline. On the aligned dataset, our model shows strong results, while on the more challenging unaligned dataset, GRF achieves a lower (better) Mean Absolute Error (MAE), indicating more precise regression predictions even when temporal alignment is not guaranteed.

\begin{table}[h!]
\centering
\begin{tabular}{@{}llcccc@{}}
\toprule
\multirow{2}{*}{\textbf{Dataset}} & \multirow{2}{*}{\textbf{Model}} & \multicolumn{4}{c}{\textbf{Metrics}} \\
\cmidrule(lr){3-6}
& & \textbf{Acc-2 $\uparrow$} & \textbf{Acc-7 $\uparrow$} & \textbf{F1-Score $\uparrow$} & \textbf{MAE $\downarrow$} \\
\midrule
\multirow{2}{*}{Aligned} & MulT Baseline & \textbf{0.830} & \textbf{0.400} & \textbf{0.828} & 0.871 \\
& GRF (ours) & 0.810 & 0.327 & 0.804 & \textbf{0.886} \\
\midrule
\multirow{2}{*}{Unaligned} & MulT Baseline & \textbf{0.811} & \textbf{0.391} & \textbf{0.810}  & 0.889 \\
& GRF (ours) & 0.796 & 0.302 & 0.794 & \textbf{0.865} \\
\bottomrule
\end{tabular}
\caption{Main performance comparison on the CMU-MOSI test set. Our GRF model is competitive with the MulT baseline.}
\label{tab:main_comparison}
\end{table}

\subsection{Ablation Study: Impact of Modality Order}
Table \ref{tab:ablation_order} details the impact of the modality fusion order. The results show that initiating the fusion process with the text modality (T) generally leads to superior outcomes, especially on the aligned dataset. This suggests that establishing a strong semantic foundation from text is a robust strategy, allowing the acoustic and visual streams to effectively modulate this anchor representation.

\begin{table}[h!]
\centering
\begin{tabular}{@{}llcccc@{}}
\toprule
\multirow{2}{*}{\textbf{Dataset}} & \multirow{2}{*}{\textbf{Fusion Order}} & \multicolumn{4}{c}{\textbf{Metrics}} \\
\cmidrule(lr){3-6}
& & \textbf{Acc-2 $\uparrow$} & \textbf{Acc-7 $\uparrow$} & \textbf{F1-Score $\uparrow$} & \textbf{MAE $\downarrow$} \\
\midrule
\multirow{3}{*}{Aligned} & A$\rightarrow$V$\rightarrow$T & 0.810 & \textbf{0.327} & 0.804 & \textbf{0.886} \\
& T$\rightarrow$V$\rightarrow$A & 0.781 & 0.307 & 0.771 & 0.936 \\
& T$\rightarrow$A$\rightarrow$V & \textbf{0.810} & 0.302 & \textbf{0.807} & 0.934 \\
\midrule
\multirow{3}{*}{Unaligned} & \textbf{A$\rightarrow$V$\rightarrow$T} & \textbf{0.796} & 0.302 & \textbf{0.794} & \textbf{0.865} \\
& T$\rightarrow$V$\rightarrow$A & 0.771 & 0.305 & 0.764 & 0.955 \\
& T$\rightarrow$A$\rightarrow$V & 0.792 & \textbf{0.327} & 0.782 & 0.947 \\
\bottomrule
\end{tabular}
\caption{Ablation on fusion order. Sequences starting with Text perform best on the aligned dataset.}
\label{tab:ablation_order}
\end{table}

\begin{figure*}[!t]
    \centering
    \includegraphics[width=\textwidth]{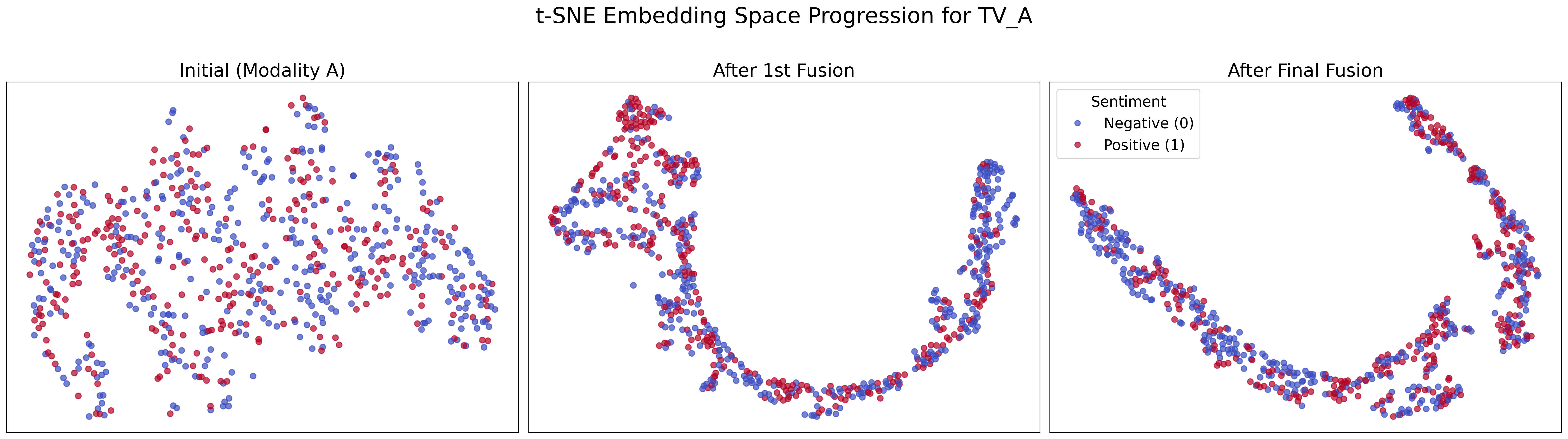}
    \vspace{-1.5ex}
    \caption{
        \textbf{t-SNE visualization of the embedding space at different stages of the fusion process.} 
        The analysis for the T$\rightarrow$V$\rightarrow$A order shows a clear progression. 
        \textbf{(Left)} The initial text representation shows no class separability. 
        \textbf{(Middle)} Fusing vision introduces a clear structure but does not separate the classes. 
        \textbf{(Right)} The final fusion with audio achieves distinct class separation, pulling positive (1) and negative (0) samples apart. This demonstrates effective, sequential multimodal integration.
    }
    \label{fig:tsne}
\end{figure*}

\subsection{Qualitative Analysis of the Fusion Process}
Beyond quantitative metrics, we visually inspect how our model learns to fuse information using t-SNE projections of the embeddings at each stage of the T$\rightarrow$V$\rightarrow$A pipeline. The results, shown in Figure~\ref{fig:tsne}, provide a clear narrative. The initial text representations (left panel) show no class separability. After fusing with the visual modality (middle panel), a distinct structure emerges, yet the sentiment classes remain intermingled. It is only after the final fusion with audio (right panel) that a clear separation between sentiment classes becomes apparent. This visual progression strongly suggests that our GRF model learns to sequentially integrate complementary information from each modality.

\begin{figure*}[!t]
    \centering
    \includegraphics[width=\textwidth]{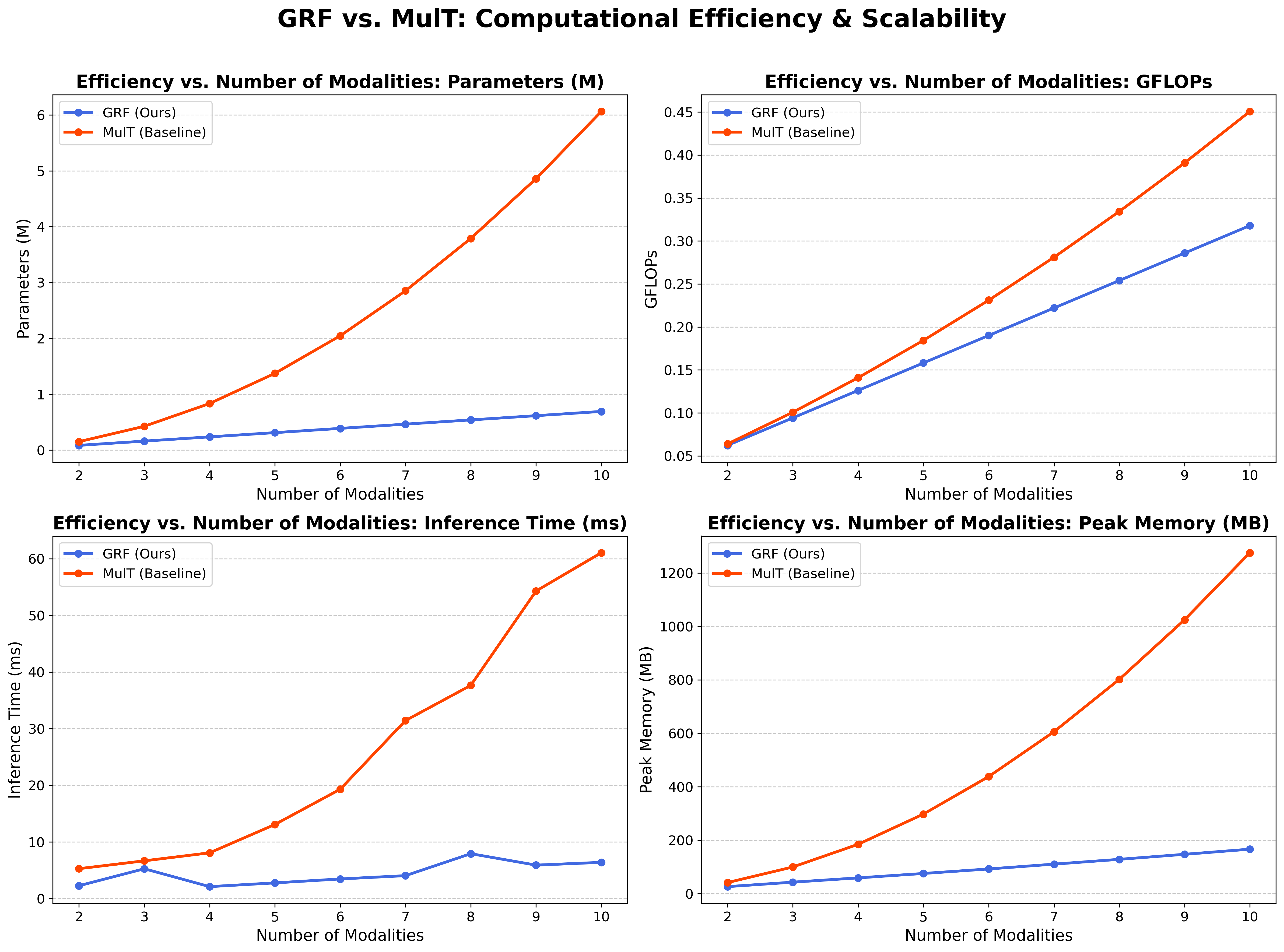}
    \vspace{-1.5ex}
    \caption{
        \textbf{Computational efficiency and scalability comparison between our sequential GRF model and the parallel MulT baseline.}
        The plots for Parameters (M) and GFLOPs clearly show the linear $\mathcal{O}(N)$ complexity of GRF versus the quadratic $\mathcal{O}(N^2)$ complexity of MulT as the number of modalities ($N$) increases. This superior theoretical efficiency translates to lower practical inference times and peak GPU memory usage, highlighting GRF's advantages for applications involving a large number of modalities.
    }
    \label{fig:efficiency}
\end{figure*}

\subsection{Computational Efficiency and Scalability}
A key advantage of our proposed sequential architecture is its computational efficiency and scalability compared to parallel pairwise approaches like MulT. To quantify this, we benchmark both models on four key metrics while varying the number of input modalities from 2 to 10. The results are presented in Figure~\ref{fig:efficiency}.

The analysis reveals a stark difference in scalability. The computational requirements of our GRF model—in terms of parameters, GFLOPs, inference time, and peak memory—grow linearly, $\mathcal{O}(n)$, with the number of modalities $n$. In contrast, the MulT baseline, which requires a cross-modal transformer for every pair of modalities, exhibits a quadratic growth rate, $\mathcal{O}(n^2)$. This inherent architectural efficiency makes our GRF model a significantly more practical and scalable solution for real-world applications that may involve more than the standard three modalities.
\section{Discussion}

Our experiments on the CMU-MOSI benchmark demonstrate that Gated Recurrent Fusion (GRF) is an effective and highly efficient architecture for multimodal learning. The central finding of our work is that by adopting a sequential, stateful fusion mechanism, we can achieve performance competitive with more complex, quadratic-cost baselines like MulT, while offering a clear and significant advantage in computational scalability. This addresses a critical need for efficient models in an era where the number of integrated modalities is poised to grow.

A key insight from our work comes from the qualitative analysis of the fusion process. The t-SNE visualizations in \Cref{fig:tsne} reveal that our GRF architecture successfully learns to \textbf{progressively refine} the embedding space. With each modality's integration, the representation becomes more structured, leading to a final state with clear class separability. This visual evidence supports our core architectural hypothesis: that a recurrent pipeline, guided by a sophisticated fusion block, can untangle and integrate complex multimodal signals in a step-by-step manner.

The design of our fusion block, which includes a GRU-inspired Gated Fusion Unit (GFU), is motivated by the need to control information flow during these sequential updates. While simple operations like addition or concatenation are viable, they lack an explicit mechanism to prevent valuable context from being diluted or overwritten by noisy or redundant features from new modalities. The GFU is theoretically designed to mitigate this by acting as a learned gate, and the strong overall performance of the GRF framework suggests that such a control mechanism is beneficial.

\paragraph{Limitations and Future Work.}
Our analysis also underscores the limitations of using a single, relatively small benchmark like MOSI. The rapid convergence we observed suggests that the dataset may not be sufficiently complex to fully differentiate the performance ceilings of advanced architectures. Therefore, the most critical next step is to evaluate GRF on larger, more challenging datasets like MOSEI or MELD, where we hypothesize that the benefits of its efficient and controlled fusion will be even more pronounced.

Furthermore, this work lays the foundation for several exciting avenues. A crucial piece of future work is to conduct a detailed \textbf{ablation study on the Gated Fusion Unit (GFU)} to empirically quantify its contribution against simpler fusion methods. Additionally, the sequential nature of GRF invites exploration into dynamic modality routing, where a model could learn the optimal fusion order for a given task.

\section{Conclusion}

In this work, we introduced Gated Recurrent Fusion (GRF), a novel architecture designed to resolve the long-standing trade-off between fusion expressiveness and computational scalability in multimodal learning. By processing modalities sequentially through a stateful pipeline, GRF leverages a powerful, decoder-inspired cross-attention mechanism while maintaining linear complexity, $\mathcal{O}(n)$, with respect to the number of modalities. This stands in stark contrast to traditional pairwise models that incur quadratic overhead.

Our empirical results on the CMU-MOSI benchmark validate our approach, demonstrating that GRF achieves competitive performance compared to established, computationally intensive baselines. Qualitative analysis through t-SNE visualizations provides strong evidence that our architecture successfully learns to progressively refine its internal representation, leading to a well-structured and class-separable embedding space.

By presenting an architecture that is both efficient by design and empirically effective, GRF offers a robust and practical paradigm for building the next generation of scalable multimodal systems. We believe it is a promising step towards deploying sophisticated fusion models in real-world, high-modality applications.

\bibliographystyle{plainnat}

\end{document}